# Optimizing Speech-Input Length for Speaker-Independent Depression Classification

*Tomasz Rutowski, Amir Harati, Yang Lu, Elizabeth Shriberg*

Ellipsis Health, Inc.

{tomek,amir,yang,liz}@ellipsishealth.com

## Abstract

Machine learning models for speech-based depression classification offer promise for health care applications. Despite growing work on depression classification, little is understood about how the length of speech-input impacts model performance. We analyze results for speaker-independent depression classification using a corpus of over 1400 hours of speech from a human-machine health screening application. We examine performance as a function of response input length for two NLP systems that differ in overall performance.

Results for both systems show that performance depends on natural length, elapsed length, and ordering of the response within a session. Systems share a minimum length threshold, but differ in a response saturation threshold, with the latter higher for the better system. At saturation it is better to pose a new question to the speaker, than to continue the current response. These and additional reported results suggest how applications can be better designed to both elicit and process optimal input lengths for depression classification.

## 1. Introduction

Depression is a prevalent disabling condition and a major global public health concern [1], [2]. Mobile AI technology could play an important role in expanding screening for depression, especially as an aid to providers who could follow up with appropriate care. Speech technology offers promise because speaking is natural, can be used at a distance, requires no special training, and carries information about a speaker's state. A growing line of AI research has shown that depression can be detected from speech signals using natural language processing (NLP), acoustic models, and multimodal models [3], [4], [5], [6], [7], [8], [9], [10]. Common evaluations with shared data sets, features, and tools have recently led to progress, especially in modeling methods [11], [12], [13], [14], [15].

With a few exceptions [16], [18] little is known about how much speech is needed from patients to get good classification performance. At a first order, "more speech is better" for evaluation in most speech technology tasks [15], [19]. But from a practical standpoint, longer inputs add time for patients and increase infrastructure costs for systems.

We investigate how input length affects classification performance in a large data set from a depression screening application. We examine length both at the level of individual patient turns or "responses", and at the level of a multi-response "session". Since better models are better able to take advantage of extra length, we also compare results across two systems that differ in overall performance.

## 2. Method

### 2.1. Data

It was necessary to use a new corpus to obtain enough data to study length effects. To facilitate comparative research we have initiated discussion with the Linguistic Data Consortium on future release of data from this corpus to the community [20]. Corpus statistics are given in Table 1. For training we used a *larger* (1400 speech hours, 9,600 unique users) set as well as a *smaller* (650 speech hours, 6,600 unique users) subset of the same data. The latter was used to create a degraded system for comparison of threshold values. Importantly, test data was held constant over both systems. Train and test partitions contain no overlapping speakers.

The data comprise American English spontaneous speech, with users allowed to talk freely [21] in response to questions within a session. Users range in age from 18 to over 65, with a mean of roughly 30. They interacted with a software application that presented questions on different topics, such as "work" or "home life". Responses average about 125 words—longer than some reports of turn lengths in conversation, e.g. [22]; see Figure 1. Users responded to 4-6 (mean 4.52) different questions per session, and then completed a PHQ-9 [23] after the suicidality question was removed. The resulting session-level PHQ-8 score served as the gold standard for both the session and the responses within it. Scores were mapped to a binary classification task, with scores at or above 10 mapped to depressed (+dep) and scores below 10 to nondepressed (-dep), following [24].

Table 1: *Corpus statistics by partition and class.*

|  | Total | Train -dep | Train +dep | Test -dep | Test +dep |
|---|---|---|---|---|---|
| ***Smaller** (650h)* | | | | | |
| Responses | 32,078 | 12,966 | 4,602 | 11,366 | 3,144 |
| Sessions | 6,794 | 2,743 | 966 | 2,430 | 655 |
| ***Larger** (1400h)* | | | | | |
| Responses | 64,518 | 35,715 | 14,293 | 11,366 | 3,144 |
| Sessions | 14,262 | 7,990 | 3,187 | 2,430 | 655 |

As shown in Figure 1, data partitions are well matched, including nearly identical CDFs (black lines overlap) and similar distributions for class lengths both within and across partitions. Depression priors (i.e. +dep) differ slightly, at 28% (*smaller* 26%) for train vs. 22% for test data.

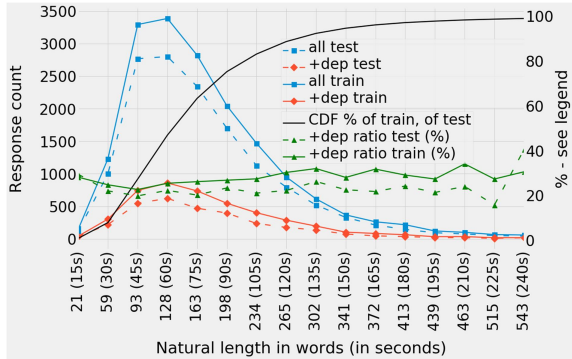

Figure 1: *Distribution of lengths in the smaller corpus.*

### 2.2. NLP Systems

Two different NLP systems were used, designed to differ in overall performance in order to test the effect on length thresholds. Both used the output of speech transcription from google *async* ASR [26]. System 1, our purposely weaker system, used an SVM and the *smaller* training set described earlier. Various word embedding techniques including Word2Vec [27], Glove [28], and ELMo [29] were investigated. In addition, different approaches for combining word vectors were tested including average, power-means [33], and z-normalization. Based on results we used Word2Vec and averaging. System 2, our purposely stronger system, used deep learning based on ULMFiT [30], [31] and the *larger* data set. Instead of using embeddings for word representations (e.g.Word2Vec), ULMFiT is a deep learning model trained on large publicly available corpora (e.g. Wikipedia data). The trained network serves as a multipurpose RNN-LSTM language model, which is fine-tuned for classification purposes. In our case we used this model to predict depression class. We are exploring updated approaches such as [32]. The number of tokens in System 1 was 7,000; in System 2 this number increased to 30,000.

## 3. Results & Discussion

### 3.1. Speaking rate

We first looked at how factors of interest (class and length) correlate with rate itself. The simplest assumption is that on average, there is a fixed relationship between words and time, for all speakers. While close, this is not the case. One reason is a lower speaking rate associated with depression, as has been previously reported in the literature (see e.g. [34]). We do find this effect, and it is ordered with depression severity as shown in Figure 2. It is, however, a small effect. The difference between the two classes is on the order of only about 5 words per minute.

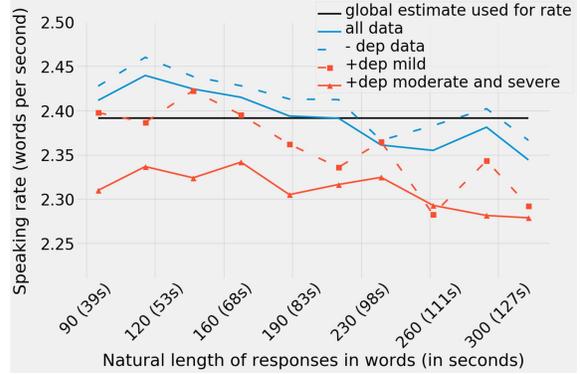

Figure 2: *Speaking rate by class and length.*

Second, there is a decline for all four curves across speaking rate, corresponding to a slight slowing for longer responses in general across classes. Thus, the longer a response is naturally, the fewer words per second it generally contains. Here the effect is on the order of about 3 or 4 words per minute. Because these effects are small, we compute a single aggregate rate over all data, of 2.39 words/second (143.4 words/minute), with which to estimate time bins for all future analyses that use only word information. The value is indicated by the dark line, or "global estimate for rate" and used in later figures to convert words to seconds.

### 3.2. Aggregate length effects

A simple way to manage a time budget would be to allow the user to speak until some target amount of total speech is reached. Figure 3 shows results for this approach. Classification systems are NLP Systems 1 and 2. Performance is reported as AUC (area under the curve), since we use a binary task, have a skewed class distribution, and have no *a priori* difference in error costs.

Length is presented using a gating measure, showing how much information is present "so far" at any point. We define the metric as follows. *Cumulative gated length* is the value of $x$ at which all data in the condition are accumulated, and at which the value of $y$ is computed after removing any additional length for data samples longer than $x$. We first examine the session performance as a function of cumulative gated length concatenated over multiple responses. Results are shown in Figure 3. For comparison, response-level results are also shown.

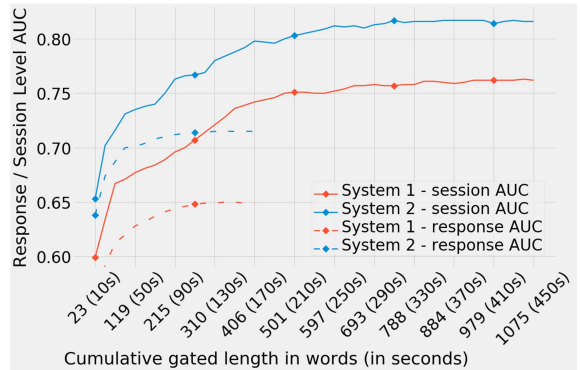

Figure 3: *AUC for sessions and utterances, using Systems 1 and 2.*

As shown, and as expected, System 2 outperforms System 1, and sessions (which concatenate all response data) outperform individual responses. (Curves stop where there is not enough data to evaluate potential additional gain.) Important observations for length include:

1. Both systems show sharp decline below 30 to 50 words.
2. Responses saturate in AUC at about 250 words.
3. Sessions appear to saturate at closer to 1000 words.

To understand the contribution of responses as they accumulate within a session, see Figure 4. As noted earlier, our data contain a mean of roughly 4.5 responses per session.

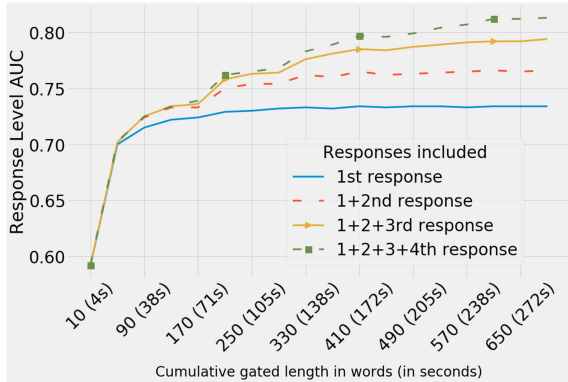

Figure 4: *Session-level AUC as a function of progressive addition of responses. Data shown for System 2 only.*

We note five observations, which apply to both Systems:

4. The session minimum is 30 to 50 words, regardless of number of responses.
5. The benefit of N+1 over N diminishes as N increases.
6. For a given length, having more responses is better; this is true for both Systems.
7. Starting a new response gives max gain of about 4%.
8. Early responses saturate at 200 words for System 2 (and at 120 words for System 1, not shown).

Point 4 reflects that even given multiple responses, NLP requires at least 30 to 50 words in order to perform. Additional responses add progressively less value as the magnitude of base increases; this is expected mathematically. Observation 6 notes it is better to compose a session of multiple responses than fewer longer ones. The largest benefit in moving to a new response is about 4% in AUC, right after the 1st response (points 5 and 6 combined). Point 8 suggests that once a response reaches saturation length, it is better to move on to a new question.

To convey performance in specificity and sensitivity, two metrics useful in the health domain, Figure 5 shows ROC results for sample gating values.

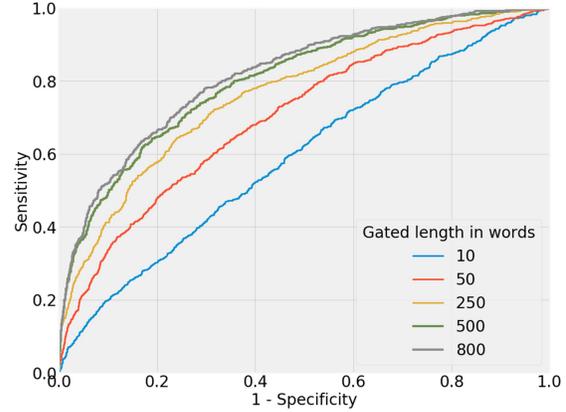

Figure 5: *System 2; Session-level performance for combined model as a function of gated session length in words.*

We note that both System 1 and System 2 outperform unaided primary care physicians as reported in [25] at 87% specificity/54% sensitivity. In addition to showing specificity and sensitivity tradeoffs, Figure 5, along with Figure 3 and Figure 4, suggests that session performance continues to improve beyond 800+ words. This suggests:

9. System 2 session length saturation is likely to be at about 1000 words, or just over 8 minutes.

### 3.3. Within-session length effects

The prior length effects were computed over all speakers and sessions. We asked whether effects are present even when controlling for the speaker and the particular session. We first asked about length ordering within a session. Since users could select which question to answer at each new utterance, we assumed they would speak more about earlier (preferred) questions. This pattern was also predicted based on a fatigue hypothesis, i.e. that speakers tire over a session, causing shortening effects. When we examined the data, however, we observed the opposite effect. We looked at the subset of sessions that had four responses. Within each session, a user's responses were ordered from longest (darkest color) to shortest (lightest color) in words. We then looked at which lengths occurred at which ordering positions. Figure 6 displays a set of bars for each of these ordering slots, 1st through 4th. Each category of bars sums to 1 e.g. "shortest". The highest bar in each set indicates the most frequent length for that ordering slot.

There is a clear predominance of shortest responses in the first position. In each subsequent position slot, the most frequent length matches the order of the position slot. We conclude from this pattern that:

10. Speakers tend to *increase* their response lengths as they progress through a session.

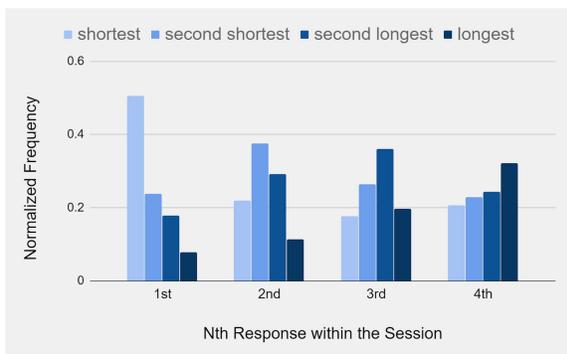

Figure 6: *Within-session length orderings for sessions with four responses. N=6300 (smaller data set, results similar for larger set).*

This effect is not explained by the questions themselves, which were patterned in many different orders of users. It is also inconsistent with a "gaming the system" hypothesis, since if speakers began with short responses, they could have continued that way. Our current hypothesis is that speakers warmed up to the system and/or the task over time.

Next, for each session we selected the shortest and the longest response (in words).

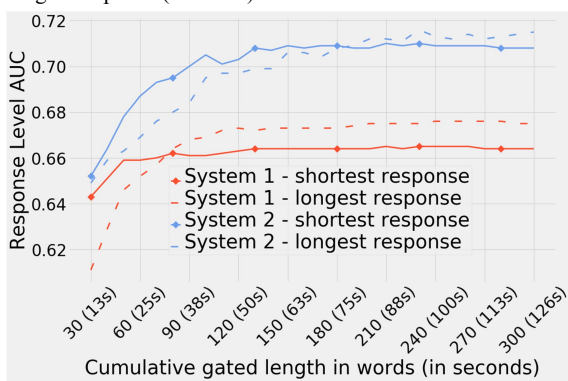

Figure 7: *AUC for shortest versus longest responses within sessions.*

We expected at least similar value from similar lengths across longer and shorter responses. Figure 7 however, shows that even when we control for the speaker and session, shorter responses outperform longer ones *initially*, while longer responses outperform shorter ones *eventually*.

Finally, we looked at where response performance begins to saturate. We took all data from a frequently-occurring natural length bin of 150 and 200 words. Each response in the bin was cut at various lengths, based on word count. Performance was then compared within a response, between the early part and later part.

Results showed that the first part of responses (60% of the total length) is less valuable than the second part by 6% on the AUC scale for System 2. We also performed the same analysis for bins of different natural lengths. In the case of a bin from 60 to 90 words, the effect was consistent; i.e. the second half was more valuable than the first. We searched empirically for natural length values of the transition in behavior from better performing second halves to better performing first halves. Results can be summarized as follows:

11. Long utterances perform better than short ones—eventually. This was consistent for both Systems.
12. Short utterances perform better than long, initially.
13. There is a threshold length below which one should not cut off a current response. This length is about 80 words for System 1, and 150 words for System 2.
14. There is a saturation length after which one should consider cutting off a response and moving to a new question. This length is after about 120 words for System 1, and about 200 words for System 2.

## 4. Summary and Conclusions

The length of speech input has clear consequences for NLP performance in depression classification. While more speech is better, practical constraints encourage optimizing length to minimize costs for both users and systems. The relationship between words and time depends on depression class as well as on natural response length. An average rate (2.39 words/second) worked for mapping one metric to the other.

Results compared two systems that differ in performance overall, to test for similarity in patterns and difference in absolute thresholds. Analyses using AUC indicate that responses for both systems should be at least 30 to 50 words long (about 20 seconds). Within a single response, there is a threshold below which one should keep waiting, and one at which it is better to move to a new question. These values depend on the system itself, with the better system making better use of additional words (80 and 120 words for System 1, respectively) versus 150 and 200 words for System 2. When interested in overall session performance, concatenating a larger number of shorter responses is better than using a smaller number of longer responses—as long as all responses exceed the minimum length. Moving to a new response provides maximum relative gain when early in a session. Using our better system, session length saturation appears to occur after about 8 minutes of total speech.

A surprising finding was that speakers tend to produce shortest lengths early in a session and speak progressively longer with each new response. This behaviour facilitates collecting more responses per session. Within-speaker analysis also showed that while longer responses perform better overall, once the utterances are completed, shorter responses perform better initially. In comparing the two systems, we confirm that while minimum thresholds are similar, the better system can make better use of additional words, resulting in higher saturation thresholds.

We expect that specific lengths and speaking rates will vary for different data sets, for example different languages, age groups, tasks, and so forth. For future systems, these findings can be optimized for new corpora to improve both data elicitation and machine learning. Elicitation systems could let users know when it is okay to proceed to a new question, to end a session, or when to provide additional speech. Machine learning can optimize efficiency by prioritizing the use of length regions that contain maximum information for classification.

## 5. Acknowledgements

We thank David Lin, Ricardo Oliveiro, Mike Aratow and Mainul Mondal for support and contributions.